\pdfoutput=1

\documentclass[11pt]{article}

\usepackage[preprint]{acl}

\usepackage{times}
\usepackage{latexsym}
\usepackage[T1]{fontenc}
\usepackage[utf8]{inputenc}
\usepackage{microtype}
\usepackage{inconsolata}
\usepackage{graphicx}
\usepackage{tabularx}
\usepackage{multirow}
\usepackage{easy-todo}
\usepackage{comment}
\usepackage{arydshln}
\usepackage{soul}
\usepackage{enumitem}
\usepackage{booktabs}
\usepackage{makecell}
\usepackage{bm}
\usepackage[dvipsnames]{xcolor}
\usepackage{pgf}
\usepackage{placeins}
\usepackage{float}
\usepackage{amssymb}
\usepackage{CJKutf8}

\usepackage{pgfplots}
\pgfplotsset{compat=1.18} % PGFPlots version
\usepgfplotslibrary{groupplots} % read groupplot
\usepackage{tikz}
\usetikzlibrary{shapes.arrows}
\usetikzlibrary {arrows.meta}
\usetikzlibrary{matrix}

\title{Towards Trustworthy Lexical Simplification: Exploring Safety and Efficiency with Small LLMs}

\author{Akio Hayakawa \And
  Stefan Bott \\
  Department of Engineering, Universitat Pompeu Fabra\\
  Barcelona, Spain\\
  \texttt{\{akio.hayakawa, stefan.bott, horacio.saggion\}@upf.edu}
  \And Horacio Saggion\\
}

\begin{document}
\maketitle

\begin{abstract}

Despite their strong performance, large language models (LLMs) face challenges in real-world application of lexical simplification (LS), particularly in privacy-sensitive and resource-constrained environments.
Moreover, since vulnerable user groups (e.g., people with disabilities) are one of the key target groups of this technology, it is crucial to ensure the safety and correctness of the output of LS systems.
To address these issues, we propose an efficient framework for LS systems that utilizes small LLMs deployable in local environments. Within this framework, we explore knowledge distillation with synthesized data and in-context learning as baselines.
Our experiments in five languages evaluate model outputs both automatically and manually. Our manual analysis reveals that while knowledge distillation boosts automatic metric scores, it also introduces a safety trade-off by increasing harmful simplifications.
Importantly, we find that the model's output probability is a useful signal for detecting harmful simplifications. Leveraging this, we propose a filtering strategy that suppresses harmful simplifications while largely preserving beneficial ones.
This work establishes a benchmark for efficient and safe LS with small LLMs. It highlights the key trade-offs between performance, efficiency, and safety, and demonstrates a promising approach for safe real-world deployment. \footnote{This paper was accepted to INLG 2025.}

\end{abstract}

\section{Introduction}

Text Simplification (TS) aims to make texts more accessible by rewriting them in simpler language. TS holds the potential to alleviate reading and understanding difficulties, particularly for individuals with dyslexia \citep{rello2013simplify}, intellectual disabilities \citep{sauberli2024digital}, and Deaf and hard-of-hearing adults \citep{alonzo2021comparison}.
TS is a task strongly oriented towards real-world scenarios, aiming to promote social participation and inclusion among people who face challenges in text comprehension.

Recent advancements in large language models (LLMs) have revolutionized natural language processing and achieved state-of-the-art performance across various tasks \citep{openai2024gpt4technicalreport}. TS is no exception, as LLMs have outperformed existing TS systems \citep{feng2023sentence, wu2024depth, qiang2025redefining}.

However, applying LLMs to TS in real-world scenarios, particularly for vulnerable user groups, faces critical challenges. First, prompts provided to LLMs and texts requiring simplification may contain sensitive personal information, such as data related to cognitive impairments. The use of API-based LLMs involves transmitting that sensitive data over the internet, raising significant privacy concerns. For instance, given that individuals with dyslexia often hesitate to disclose their condition due to concerns about stigma and negative perceptions \citep{hamilton2024dyslexia}, it can be problematic to design prompts for TS such as \textit{"I have dyslexia; Can you simplify this diagnosis result for me?"}. Thus, TS systems capable of running locally are highly desirable.

Open-access LLMs address this privacy concern. However, high-performing open-access LLMs typically require substantial computational resources for inference. Deploying such large models directly on resource-constrained devices, such as smartphones and tablets that are commonly used by the target users \citep{soderstrom2021using}, is currently impractical. This highlights the need for developing smaller models that can perform effectively within these limited hardware environments. 

Building on these challenges, we investigate how to develop efficient TS systems that can operate under constrained computational resources. This approach is essential for supporting information access for all while respecting user privacy.

Utilizing small LLMs is a promising approach, as \textasciitilde3B models are often explicitly engineered for on-device deployment \citep{metallama32}, thereby addressing privacy and efficiency issues.
However, particular attention must be paid to safety when employing small LLMs, as their limited capacity compared to larger counterparts introduces critical considerations regarding the reliability and harmfulness of the generated simplifications.
Poor or inaccurate simplifications can be detrimental, as they may actively provide misinformation or cause confusion, which are more serious issues than leaving the text unchanged \citep{rello2013simplify, sauberli2024digital}.
Therefore, in practice, it is crucial not only to simplify texts effectively, but also to minimize harmful outputs and ensure safety.

As a first step towards addressing these challenges, this paper focuses specifically on lexical simplification (LS), a subtask of TS that replaces complex words in a context sentence with simpler alternatives. LS can be considered a relatively conservative and safe subtask compared to sentence- or document-level simplification, which often involves operations such as information deletion \citep{al2021automated}.

We adopted small LLMs and explored two approaches: in-context learning, which requires no training, and knowledge distillation, which transfers knowledge from a larger teacher model to a smaller student model.
Our approach also considers extensibility to diverse languages, as supporting a broad user group requires simplification across multiple languages.

To evaluate the safety of simplification outputs, particularly in suppressing harmful content, we conducted manual evaluations alongside automatic metrics. Manual analysis revealed that, while knowledge distillation generally boosted automatic metric scores, it did not reduce harmful outputs and sometimes even increased them. Furthermore, we observed that, especially in models trained via knowledge distillation, the output probability provided by LLMs may serve as a useful signal for identifying harmful simplifications.\footnote{Our codes will be available at \url{https://github.com/ahaya3776/safe-efficient-ls}.}

Our contributions are summarized as follows:
\begin{itemize}[itemsep=0.2em, parsep=0em]
    \item We investigated the potential and challenges of using small LLMs for lexical simplification with respect to safety and efficiency, and we establish a benchmark in this important research area.
    \item We demonstrated that small LLMs offer significant inference speedups, which highlights their efficiency.
    \item We found that standard approaches such as in-context learning and knowledge distillation can produce beneficial simplifications, but they inherently risk generating harmful outputs. 
    \item We identified that model's log-probability serves as a useful signal for detecting harmful simplifications, suggesting a promising filtering strategy to ensure safety towards real-world applications.
\end{itemize}

\section{Related Work}

\paragraph{Lexical Simplification}
LSBert \citep{qiang2021lsbert} established itself as a strong baseline for LS by leveraging BERT's unmasking capabilities and contextual understanding, outperforming earlier systems based on paraphrase databases and word embeddings \citep{biran-etal-2011-putting, glavas-stajner-2015-simplifying}. However, such systems based on masked language models (MLMs) were limited in generating multi-token words \citep{przybyla-shardlow-2020-multi} and its effectiveness outside English has been questioned \citep{stajner-etal-2023-less}. Furthermore, MLM-based systems often require multi-stage pipelines involving candidate ranking, which introduces significant latency that conflicts our goal of on-device efficiency. Their multilingual applicability is also hindered by the inconsistent availability of monolingual models across languages.

More recent auto-regressive approaches, using T5 \citep{sheang-saggion-2021-controllable} and GPT-3 \citep{aumiller-gertz-2022-unihd}, have outperformed MLM-based methods, leading to the widespread adoption of LLMs as the predominant solution for LS \citep{shardlow-etal-2024-bea}. Notably, a GPT-4-based LS system \citep{enomoto-etal-2024-tmu} achieved remarkable performance across multiple languages.

\paragraph{Smaller LLMs and Efficiency}

The use of high-performing versatile LLMs poses several challenges in real-world scenarios, including resource limitations, privacy concerns, and high operational costs.
To address these issues, various efforts have been made to develop LLMs capable of running on local devices. These include techniques such as quantization \citep{zhou2024surveyefficientinferencelarge} and the GPT-Generated Unified Format (GGUF),\footnote{\url{https://github.com/ggml-org/ggml/blob/master/docs/gguf.md}} both of which aim to enable efficient inference without high-end hardware, as well as the development of small LLMs \citep{qwen2.5, gemmateam2024gemma2improvingopen, meta2024llama3}.

Small LLMs can be further trained to improve performance on specific tasks \citep{xu2024survey}, including LS \citep{baez-saggion-2023-lsllama, xiao-etal-2024-optimizing}. \citet{baez-saggion-2023-lsllama} proposed LSLlama, a LLAMA-7B model fine-tuned on an existing LS dataset, which achieved performance comparable to a GPT-3-based approach. \citet{xiao-etal-2024-optimizing} introduced the PivotKD framework, which trained Chinese-centric small LLMs using pseudo-instances generated by GPT-4, and built a cost-effective Chinese LS system by incorporating web-based synonym and word sense retrieval during inference.
These studies demonstrated the potential of task-specific training of small LLMs for LS. However, their applicability to languages beyond English and Chinese remains uncertain, especially given morphological complexity and disparities in pre-training resources.

\paragraph{Safety and Reliability of Text Simplification}
While TS supports reading and understanding, it also carries the risk of causing confusion or misinterpretation. In practice, outputs from automatic TS systems often suffer from low factuality \citep{devaraj-etal-2022-evaluating} and information loss \citep{agrawal-carpuat-2024-text}, which can negatively affect readers' reading time and accuracy on comprehension questions \citep{rello2013simplify, sauberli2024digital}.
In such cases, leaving the original text unchanged may be preferable to applying a harmful simplification. Therefore, adopting a strategy that accepts simplification only when certain criteria are met offers a practical approach in real-world scenarios. In this regard, \citet{trienes-etal-2024-infolossqa} presented one of the few efforts to assess the potential harm of TS by detecting information loss. However, its reliance on LLMs makes it unsuitable for use in constrained environments.

\section{Experimental Setup}
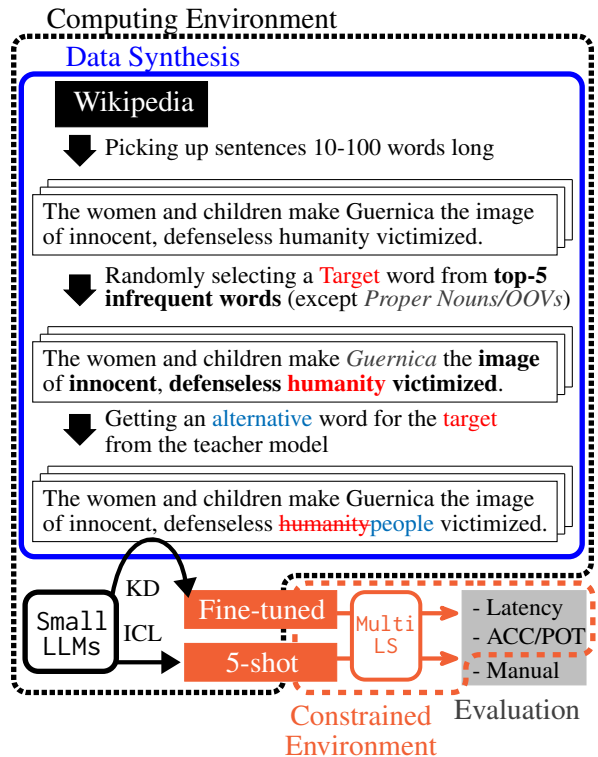
\begin{figure}[t]
\centering
\begin{tikzpicture}[scale=1.00]

\draw[blue, rounded corners=5pt, line width=2pt] (0.05, 0.70) rectangle (7.43, 7.10);
\node[above, right, blue, font=\normalsize] at (0.50, 7.30) {Data Synthesis};
\draw[black, rounded corners=5pt, line width=2pt, densely dotted] (-0.05, 7.60) -- (7.55, 7.60) -- (7.55, 0.45) -- (3.5, 0.45) -- (3.5, -1.10) -- (-0.05, -1.10) -- cycle;
\draw[RedOrange, rounded corners=5pt, line width=2pt, dashed] (6.00, 0.35) -- (3.65, 0.35) -- (3.65, -1.10) -- (5.00, -1.10);
\node[above, right, black, font=\normalsize] at (0.25, 7.80) {Computing Environment};

\filldraw[right, black] (0.5, 6.4) rectangle (2.5, 7.0);
\node[right, white] at (0.6, 6.7) {Wikipedia};

\node[single arrow, draw=black, fill=black, 
      minimum width=1pt, single arrow head extend=2pt, rotate=270, minimum height=2mm] at (0.82, 6.15) {};
\node[below, right, font=\footnotesize] at (1.02, 6.1) {Picking up sentences 10-100 words long};
      
\draw[black] (0.2, 5.5) rectangle (7.1, 4.7);
\node[below, right, font=\footnotesize] at (0.2, 5.25) {The women and children make Guernica the image};
\node[below, right, font=\footnotesize] at (0.2, 4.92) {of innocent, defenseless humanity victimized.};
\draw[black] (0.3, 5.5) -- (0.3, 5.6);
\draw[black] (0.3, 5.6) -- (7.2, 5.6);
\draw[black] (7.2, 5.6) -- (7.2, 4.8);
\draw[black] (7.2, 4.8) -- (7.1, 4.8);
\draw[black] (0.4, 5.6) -- (0.4, 5.7);
\draw[black] (0.4, 5.7) -- (7.3, 5.7);
\draw[black] (7.3, 5.7) -- (7.3, 4.9);
\draw[black] (7.3, 4.9) -- (7.2, 4.9);

\node[single arrow, draw=black, fill=black, 
      minimum width=1pt, single arrow head extend=2pt, rotate=270, minimum height=2mm] at (0.82, 4.30) {};
\node[below, right, font=\footnotesize] at (1.02, 4.40) {Randomly selecting a \textcolor{red}{Target} word from \textbf{top-5}};
\node[below, right, font=\footnotesize] at (1.02, 4.10) {\textbf{infrequent words} (except \textcolor{darkgray}{\textit{Proper Nouns/OOVs}})};

\draw[black] (0.2, 3.55) rectangle (7.1, 2.75);
\node[below, right, font=\footnotesize] at (0.2, 3.30) {The women and children make \textcolor{darkgray}{\textit{Guernica}} the \textbf{image}};
\node[below, right, font=\footnotesize] at (0.2, 2.97) {of \textbf{innocent}, \textbf{defenseless} \textcolor{red}{\textbf{humanity}} \textbf{victimized}.};
\draw[black] (0.3, 3.55) -- (0.3, 3.65);
\draw[black] (0.3, 3.65) -- (7.2, 3.65);
\draw[black] (7.2, 3.65) -- (7.2, 2.85);
\draw[black] (7.2, 2.85) -- (7.1, 2.85);
\draw[black] (0.4, 3.65) -- (0.4, 3.75);
\draw[black] (0.4, 3.75) -- (7.3, 3.75);
\draw[black] (7.3, 3.75) -- (7.3, 2.95);
\draw[black] (7.3, 2.95) -- (7.2, 2.95);

\node[single arrow, draw=black, fill=black, 
      minimum width=1pt, single arrow head extend=2pt, rotate=270, minimum height=2mm] at (0.82, 2.45) {};
\node[below, right, font=\footnotesize] at (1.02, 2.50) {Getting an \textcolor{NavyBlue}{alternative} word for the \textcolor{red}{target}};
\node[below, right, font=\footnotesize] at (1.02, 2.20) {from the teacher model};

\draw[black] (0.2, 1.70) rectangle (7.1, 0.90);
\node[below, right, font=\footnotesize] at (0.2, 1.45) {The women and children make Guernica the image};
\node[below, right, font=\footnotesize] at (0.2, 1.12) {of innocent, defenseless \textcolor{red}{\st{humanity}}\textcolor{NavyBlue}{people} victimized.};
\draw[black] (0.3, 1.70) -- (0.3, 1.80);
\draw[black] (0.3, 1.80) -- (7.2, 1.80);
\draw[black] (7.2, 1.80) -- (7.2, 1.00);
\draw[black] (7.2, 1.00) -- (7.1, 1.00);
\draw[black] (0.4, 1.80) -- (0.4, 1.90);
\draw[black] (0.4, 1.90) -- (7.3, 1.90);
\draw[black] (7.3, 1.90) -- (7.3, 1.10);
\draw[black] (7.3, 1.10) -- (7.2, 1.10);

\draw[black, rounded corners=5pt, line width=2pt] (0.10, -0.82) rectangle (1.3, 0.23);
\node[below, font=\normalsize] at (0.72, 0.12) {\texttt{Small}};
\node[below, font=\normalsize] at (0.72, -0.22) {\texttt{LLMs}};
\node[below, RedOrange, font=\normalsize] at (4.52, -1.13) {Constrained};
\node[below, RedOrange, font=\normalsize] at (4.52, -1.53) {Environment};
\filldraw[RedOrange] (2.20, -0.25) rectangle (4.2, 0.25);
\filldraw[RedOrange] (2.20, -0.95) rectangle (4.2, -0.45);
\node[white] at (3.22, 0.0) {Fine-tuned};
\node[white] at (3.22, -0.7) {5-shot};

\draw[ultra thick, arrows={-Triangle[]}] (1.25, 0.20) to [in=100, out=80, looseness=2.5] (2.25, 0.15);
\node[above, font=\footnotesize] at (1.65, 0.05) {KD};
\draw[ultra thick, arrows={-Triangle[]}] (1.30, -0.7) to (2.15, -0.7);
\node[above, font=\footnotesize] at (1.65, -0.55) {ICL};

\draw[RedOrange, rounded corners=5pt, line width=1.5pt] (4.4, 0.23) rectangle (5.3, -0.93);
\node[RedOrange, font=\footnotesize] at (4.85, -0.18) {\texttt{Multi}};
\node[RedOrange, font=\footnotesize] at (4.85, -0.48) {\texttt{LS}};
\draw[ultra thick, RedOrange] (4.20, -0.05) to (4.4, -0.05);
\draw[ultra thick, RedOrange] (4.20, -0.65) to (4.4, -0.65);
\draw[ultra thick, RedOrange, arrows={-Stealth[round]}] (5.30, -0.05) to (5.80, -0.05);
\draw[ultra thick, RedOrange, arrows={-Stealth[round]}] (5.30, -0.65) to (5.80, -0.65);
\filldraw[lightgray] (5.85, 0.25) rectangle (7.48, -1.00);
\node[black, right, font=\footnotesize] at (5.85, 0.0) {- Latency};
\node[black, right, font=\footnotesize] at (5.85, -0.35) {- ACC/POT};
\node[black, right, font=\footnotesize] at (5.85, -0.8) {- Manual};
\draw[RedOrange, rounded corners=5pt, line width=2pt, dashed] (6.05, 0.35) -- (7.57, 0.35) -- (7.57, -0.6) -- (5.90, -0.6) -- (5.90, -1.10) -- (5.05, -1.10);
\node[below, left, darkgray, font=\normalsize] at (7.54, -1.33) {Evaluation};

\end{tikzpicture}
\caption{Overall flow of our experiments. We developed and evaluated systems for each language separately.}
\label{fig:flow}
\end{figure}

\autoref{fig:flow} illustrates the overall flow of our system development and evaluation. 
We used the HuggingFace Transformers library\footnote{\url{https://huggingface.co/docs/transformers/}} for the development of our LS models. A single Tesla T4 GPU with 16 GB of memory was used for the development. To enable high-speed inference on CPUs, the models were converted into the GGUF format using llama.cpp.\footnote{\url{https://github.com/ggml-org/llama.cpp}}

\subsection{Task Formulation}
The term \textit{Lexical Simplification} (LS) has been used with varying scopes. In some cases, it refers to a sentence-level simplification pipeline consisting of complex word identification, substitution generation, and ranking \citep{paetzold2017survey}.
However, in this paper, we adopt a narrower definition of LS, focusing solely on the substitution generation. Specifically, we define LS as generating a simpler alternative to a single target word that appears in a given context sentence.
An alternative should make the context easier to understand than the original while preserving its meaning. Therefore, an LS system takes a context and target word as input and outputs a single alternative word.

\subsection{Dataset}
\label{sec:dataset}
We used MultiLS \citep{shardlow-etal-2024-multilingual}, a LS dataset covering 10 languages, to evaluate system performance.
We selected five languages, English, Spanish, Catalan, German, and Japanese, to account for differences in language family, morphological structure, and resource availability.

\autoref{tab:ex-multils} shows an example LS instance, consisting of a context sentence, a target word, and alternative words suggested by multiple human annotators.
MultiLS allowed annotators to use a target as an alternative when they could not identify a valid simplification, which often occured when the target was already simple enough \citep{shardlow-etal-2024-extensible}.
This annotation scheme enables us to exclude instances where LS is inherently difficult. We removed such instances where the top-ranked alternative was unchanged from the target word. This process resulted in the number of instances per language shown in \autoref{tab:num-instances}.
We randomly split the selected instances into two parts, assigning 90 instances for development and the rest for testing.\footnote{As up to three instances share the same context, we assign 90 instances with 30 unique contexts to the development data.}

\subsection{LS Systems}
We employed two small LLMs: Qwen 2.5 1.5B (Qwen for short) \citep{qwen2.5} and Llama 3.2 1B (Llama for short) \citep{meta2024llama3}. Both models were trained on multiple languages from their larger counterparts.\footnote{We used base LLMs instead of instruction-tuned versions as base LLMs. See \autoref{sec:it-models} for details.} To make these models perform LS, we adopted two approaches: in-context learning and knowledge distillation.\footnote{See \autoref{sec:hps} for the hyperparameter settings.}

\subsubsection{In-Context Learning}
\label{sec:few-shot}
In-context learning \citep{brown2020languagemodelsfewshotlearners}, which provides several examples as few-shot to guide model behavior, is a common technique to improve output quality.
We used five fixed examples in the prompt \textbf{(5-shot)} following the template in \autoref{sec:prompts}. These examples were sampled from the pilot split of MultiLS, which was separated from the main evaluation data.

\subsubsection{Knowledge Distillation}
Knowledge distillation, which involves transferring knowledge of larger teacher models to smaller student models, has been widely used to adapt LLMs to specific tasks, including LS \citep{baez-saggion-2023-lsllama, xiao-etal-2024-optimizing}. Recent approaches commonly employ simple supervised fine-tuning of student models with hard labels derived from teacher model outputs, due to the advanced capabilities of closed-source LLMs \citep{xu2024survey}.
Following this framework, we performed knowledge distillation (\textbf{fine-tuned}) by synthesizing LS instances.

\paragraph{Synthesizing Context and Targets}
We randomly extracted context sentences from Wikipedia for each language. Sentences were parsed using MeCab\footnote{\url{https://taku910.github.io/mecab/}} for Japanese and spaCy\footnote{\url{https://spacy.io/}} for the other languages. We retained only those containing between 10 and 100 words as contexts.\footnote{For Japanese, simple tokenization rules were applied. See \autoref{sec:ja-tokens} for details.}

To ensure that target words were simplifiable, we excluded proper nouns and out-of-vocabulary words from the set of candidate words within each context sentence. From the remaining candidates, we randomly selected one of the five least frequent words as the target word, based on Zipf frequency.\footnote{Calculated using wordfreq Python library: \url{https://github.com/rspeer/wordfreq/}}

\begin{table}[t]
    \small
    \centering
    \begin{tabular}{p{7cm}}
    \toprule
    \textbf{Context}: Electronically controlled motorized zoom lenses are placed on both camera and projector, and synchronized with one another so that both lenses zoom together and at the same \underline{focal} length at all times.\\
    \textbf{Target Word}: focal\\
    \textbf{Gold Alternatives}: main, main, central, central, basic, primary, \st{focal}\\
    \bottomrule
    \end{tabular}
    \caption{Example from the MultiLS English subset. For this instance, \textbf{ACC} is met if the output alternative is "main" or "central", which are the most suggested alternatives. \textbf{POT} is met if the output alternative is one of "main", "central", "basic", and "primary". If the output alternative is "focal", which is unchanged from the target word, it does not meet either metric.}
    \label{tab:ex-multils}
\end{table}

\begin{table}[t]
    \small
    \centering
    \begin{tabular}{cccc}
     & \# Original & \# Selected & Avg. Context \\
    Language & Instances & Instances & Length\\
    \midrule
    English & 570 & 515 & 25.4 \\
    Spanish & 593 & 502 & 29.3 \\
    Catalan & 445 & 261 & 45.0 \\
    German & 570 & 547 & 37.7 \\
    Japanese & 570 & 562 & 20.3 \\
    \bottomrule
    \end{tabular}
    \caption{Statistics of MultiLS instances per language.}
    \label{tab:num-instances}
\end{table}

\paragraph{Synthesizing Alternative Words}
To obtain alternative words for the context-target pairs described above, we employed the instruction-tuned Gemma 2 9B \citep{gemmateam2024gemma2improvingopen} as a teacher model, an LLM known for its strong performance across diverse languages.
The model was prompted to generate a single alternative word using the same 5-shot setting described in \autoref{sec:few-shot}.

The performance of fine-tuned student models can often be improved by removing low-quality outputs from the teacher \citep{jung2023impossible, huang-etal-2023-large}. Therefore, we filtered out low-confidence alternatives, approximating confidence using output probabilities (described later in \autoref{sec:probs}).
For each language, we generated alternatives for 60,000 synthesized context-target pairs and selected the top 30,000 high-confidence instances for training.

\paragraph{Fine-tuning Models}
We fine-tuned each student model for each language separately, using the corresponding 30,000 instances for up to five epochs.
To reduce memory consumption, we adopted the QLoRA framework \citep{dettmers2023qlora}. In this setup, the weights of base models were quantized to 4-bit precision using the bitsandbytes\footnote{\url{https://github.com/bitsandbytes-foundation/bitsandbytes}} library. Fine-tuning was then performed via 16-bit LoRA adapters. 
Following \citet{dettmers2023qlora}, we only fine-tuned Query and Key projections layers within the attention modules.
Each type of student model was fine-tuned with three different random seeds.
We saved a checkpoint every 0.2 epochs and selected the one that achieved the highest Potential@1 (described later in \autoref{sec:metrics}) on the development set.
The prompt template in \autoref{sec:prompts} was used for training and inference.

\subsubsection{Baselines}
As a baseline, we employed the instruction-tuned Gemma 2 9B (Gemma for short) in the same 5-shot setting used for the teacher model.

\begin{table*}[t]
    \centering
    \setlength{\tabcolsep}{4pt}
    \small
    \begin{tabular}{c|c|cccccccccc|cccc}
         \multicolumn{2}{c|}{} & \multicolumn{10}{c|}{LS Performance} & \multicolumn{4}{c}{Latency (msec / token)}\\
        \bottomrule
        \multirow{2}{*}{Model} & \multirow{2}{*}{Settings} & \multicolumn{2}{c}{English} & \multicolumn{2}{c}{Spanish} & \multicolumn{2}{c}{Catalan} & \multicolumn{2}{c}{German} & \multicolumn{2}{c|}{Japanese} & \multicolumn{2}{c}{m6g.large} & \multicolumn{2}{c}{m6g.xlarge}\\
         & & ACC & POT & ACC & POT & ACC & POT & ACC & POT & ACC & POT & read & pred & read & pred\\
         \midrule
        %\hline
        Gemma(9B) & 5-shot & .529 & .751 & .427 & .774 & .333 & .690 & .405 & .643 & .252 & .494 & 652 & 581 & 326 & 292\\
        \midrule
        \multirow{2}{*}{Qwen(1.5B)} & 5-shot & .358 & .534 & .274 & .473 & .076 & .205 & \textbf{.186} & \textbf{.298} & .064 & .150  & 91 & 275 & 45 & 139\\
         & fine-tuned & \textbf{.382} & \textbf{.574} & \textbf{.318} & \textbf{.537} & \textbf{.129} & \textbf{.265} & .119 & .206 & \textbf{.076} & \textbf{.154} & 86 & 274 & 43 & 138\\
        \midrule
        \multirow{2}{*}{Llama(1B)} & 5-shot & .202 & .278 & .053 & .092 & .047 & .105 & .090 & .142 & .023 & .042 & 70 & 219 & 35 & 110 \\
         & fine-tuned & \textbf{.370} & \textbf{.544} & \textbf{.293} & \textbf{.529} & \textbf{.160} & \textbf{.292} & \textbf{.138} & \textbf{.217} & \textbf{.058} & \textbf{.145}  & 66 & 221 & 33 & 107\\
        \bottomrule
    \end{tabular}
    \caption{Performance of models on the MultiLS dataset. Gemma was quantized to 4-bit due to memory constraints.}
    \label{tab:result_multils}
\end{table*}

\subsection{Evaluation}

\subsubsection{Automatic LS Metrics}
\label{sec:metrics}
To automatically evaluate the performance of LS systems, we used \textbf{Accuracy@1@top1 (ACC)} and \textbf{Potential@1 (POT)}, as defined in \citet{saggion-etal-2022-findings}.
As shown in \autoref{tab:ex-multils}, ACC is the percentage of predictions matching the most frequently suggested alternative. POT is the percentage of predictions matching any suggested alternative.
Given that all instances were assumed simplifiable after the selection process in \autoref{sec:few-shot}, any predictions unchanged from the target word were not considered a match for either ACC or POT, even if the target word was included in the gold alternatives.

\subsubsection{Latency Evaluation}
To estimate model response time in resource-constrained environments, we constructed a virtual small-scale infrastructure using computing instances from Amazon Web Services (AWS).
We selected m6g.large and m6g.xlarge computing instances from AWS Elastic Computing Cloud, which provide 2 and 4 virtual CPUs and 8 GB and 16 GB of memory, respectively. These configurations reflect the hardware commonly found in smartphones and tablets. Both computing instances are based on Graviton processors, which are widely applied in mobile devices.\footnote{\url{https://aws.amazon.com/ec2/instance-types/m6g/}}

Total latency mainly consists of prompt processing time and inference time. As both depend on the number of tokens in the prompt and the generated output, we measured the average pre-token prompt processing and inference times for each model using llama.cpp.
Notably, llama.cpp caches the initial fixed portion of the prompt (i.e., few-shot examples), so its processing latency is not incurred on subsequent inferences. While this caching is key to the efficiency, it makes dynamic prompting strategies impractical, as they would require frequent cache invalidation.

\subsubsection{Manual LS Evaluation}
To gain a more nuanced understanding of LS quality and safety from a user perspective, we conducted a manual evaluation. We randomly sampled 100 instances per language and assigned harmfulness tags to the alternatives generated by each system. Our manual evaluation focused on instances that were not covered by our automatic metrics. For this purpose, we only assigned tags to alternatives that were neither unchanged from the target nor included in the gold alternatives.

Taking into account the standard human evaluation criteria of fluency, adequacy, and simplicity in TS, we defined the following four harmful tags: 

\begin{itemize}[itemsep=0.2em, parsep=0em]
    \item \textit{Grammar Error}: The alternative is grammatically incorrect, including inflection, and conjugation errors.
    \item \textit{Change of Meaning}: Replacing the target with the alternative drastically changes the meaning of context.
    \item \textit{More Difficult}: The alternative is clearly more difficult than the target, even though it preserves the meaning to some extent.
    \item \textit{Gibberish}: The alternative does not make sense at all.
\end{itemize}

For each language, annotation was performed by a single in-house annotator, all of whom were native speakers except for Catalan. The Catalan annotation was conducted by a CEFR C1 level speaker with over ten years of experience. The task was designed as a simple binary decision to minimize subjectivity, ensuring the evaluation framework is easily extensible to other languages and domains.

Based on the automatically and manually assigned tags, alternatives were categorized into following three groups. Tags determined by automatic metrics are marked with \textsc{A}, while those requiring manual annotation are marked with \textsc{M}.

\begin{itemize}[leftmargin=1.3em, itemsep=0.15em, parsep=0em]
    \item \texttt{Beneficial}
        \begin{itemize}[leftmargin=1.0em, itemsep=0.2em, parsep=0em]
            \item \texttt{ACC} (\textsc{A}) : equivalent to Accuracy@1@top1.
            \item \texttt{POT} (\textsc{A}) : Potential@1 but not \texttt{ACC}
            \item \texttt{Good} (\textsc{M}) : no harmful tags were assigned.
        \end{itemize}
    \item \texttt{Unchanged} (\textsc{A}) : alternative was identical to target.
    \item \texttt{Harmful}
        \begin{itemize}[leftmargin=1.0em, itemsep=0.2em, parsep=0em]
            \item \texttt{Degraded} (\textsc{M}) : one or more non-\textit{Gibberish} harmful tags were assigned.
            \item \texttt{Gibberish} (\textsc{M}) : \textit{Gibberish} was assigned.
        \end{itemize}
\end{itemize}

See \autoref{sec:tags} for detailed examples of the harmful tags and groups.

\subsubsection{Filtering Strategy}
To address the risk of introducing harmful simplifications discussed above, we propose and evaluate a filtering strategy.
This strategy leverages the output probability score as a reliability signal in a threshold-based decision mechanism to determine whether to perform LS.

\paragraph{Probability Score}
\label{sec:probs}
We computed the probability score as the sum of the log-probabilities of the tokens forming the alternative word, including the token indicating the end of the word (e.g., a newline or EOS token). 
We considered the probability scores of individual alternatives as candidate thresholds. For each threshold value, alternatives with scores above the threshold were accepted, while others were rejected, and no simplification was applied.

\paragraph{Evaluation}
To quantitatively evaluate the effectiveness of the proposed strategy, we defined the following metrics:
\begin{itemize}[itemsep=0.2em, parsep=0em]
    \item \textbf{AUC} (\texttt{Beneficial} vs \texttt{Harmful}): To assess how well the probability score functions as a safety signal, we computed the Area Under the ROC Curve \citep{bradley1997use} for classifying alternatives as \texttt{Beneficial} vs. \texttt{Harmful}, excluding \texttt{Unchanged} alternatives.
    \item $\bm{B_{H_{0.1}}}$ (\texttt{Beneficial} Rate at 10\% \texttt{Harmful}): To quantify practical benefit under a safety constraint, we reported the rate of \texttt{Beneficial} achieved when the rate of \texttt{Harmful} introduced was limited to 10\% of total instances.
    We chose the 10\% threshold to balance safety and utility by offering a practical reference point for comparison that remains adaptable to different needs.
\end{itemize}

\definecolor{C1}{RGB}{26,100,2} % DarkGreen
\definecolor{C2}{RGB}{84,153,6} % Green
\definecolor{C3}{RGB}{255,233,50} % Yellow
\definecolor{C4}{RGB}{128,128,128} % Gray
\definecolor{C5}{RGB}{253,153,27} % Orange
\definecolor{C6}{RGB}{255,0,0} % Red

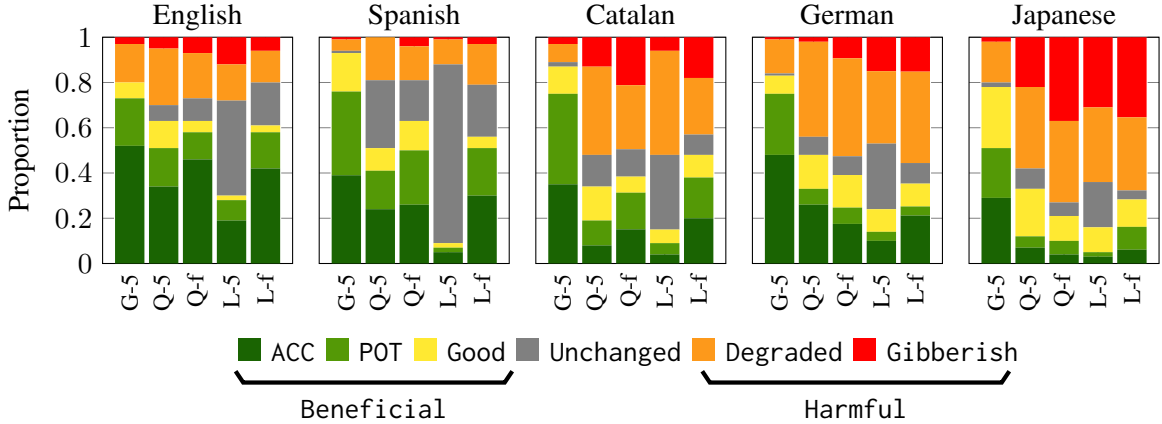
\begin{figure*}[t]
\begin{tikzpicture}[scale=1.00]

\begin{groupplot}[
    group style={
        group size=5 by 1,
        horizontal sep=0.35cm,
    },
    clip=false,
    ybar stacked,
    ymin=0,
    ymax=1.0,
    symbolic x coords={G-5, Q-5, Q-f, L-5, L-f},
    scale only axis,
    width=2.5cm,
    height=3.0cm,
    xtick=data,
    xticklabel style={
        rotate=90,
        anchor=east,
        font=\small
    },
    yticklabel style={/pgf/number format/fixed, /pgf/number format/precision=1},
    every axis y label/.style={at={(ticklabel cs:0.5)},rotate=90,anchor=near ticklabel}
]

% --- English ---
\nextgroupplot[
    title=English,
    title style={yshift=-1.5ex},
    xticklabels={G-5, Q-5, Q-f, L-5, L-f},
    enlarge x limits=0.2,
    ylabel={Proportion},
    ytick={0,0.2,0.4,0.6,0.8,1.0},
    bar width=11pt,
]
    \addplot+[ybar, fill=C1, draw=none] coordinates {(G-5,0.52)(Q-5,0.34)(Q-f,0.46)(L-5,0.19)(L-f,0.42)};
    \addplot+[ybar, fill=C2, draw=none] coordinates {(G-5,0.21)(Q-5,0.17)(Q-f,0.12)(L-5,0.09)(L-f,0.16)};
    \addplot+[ybar, fill=C3, draw=none] coordinates {(G-5,0.07)(Q-5,0.12)(Q-f,0.05)(L-5,0.02)(L-f,0.03)}; 
    \addplot+[ybar, fill=C4, draw=none] coordinates {(G-5,0.0)(Q-5,0.07)(Q-f,0.1)(L-5,0.42)(L-f,0.19)}; 
    \addplot+[ybar, fill=C5, draw=none] coordinates {(G-5,0.17)(Q-5,0.25)(Q-f,0.2)(L-5,0.16)(L-f,0.14)}; 
    \addplot+[ybar, fill=C6, draw=none] coordinates {(G-5,0.03)(Q-5,0.05)(Q-f,0.07)(L-5,0.12)(L-f,0.06)}; 

% --- Spanish ---
\nextgroupplot[
    title=Spanish,
    title style={yshift=-1.5ex},
    xticklabels={G-5, Q-5, Q-f, L-5, L-f},
    yticklabels={},
    enlarge x limits=0.2,
    bar width=11pt,
]
    \addplot+[ybar, fill=C1, draw=none] coordinates {(G-5,0.39)(Q-5,0.24)(Q-f,0.26)(L-5,0.05)(L-f,0.3)};
    \addplot+[ybar, fill=C2, draw=none] coordinates {(G-5,0.37)(Q-5,0.17)(Q-f,0.24)(L-5,0.02)(L-f,0.21)};
    \addplot+[ybar, fill=C3, draw=none] coordinates {(G-5,0.17)(Q-5,0.1)(Q-f,0.13)(L-5,0.02)(L-f,0.05)};
    \addplot+[ybar, fill=C4, draw=none] coordinates {(G-5,0.01)(Q-5,0.3)(Q-f,0.18)(L-5,0.79)(L-f,0.23)};
    \addplot+[ybar, fill=C5, draw=none] coordinates {(G-5,0.05)(Q-5,0.19)(Q-f,0.15)(L-5,0.11)(L-f,0.18)};
    \addplot+[ybar, fill=C6, draw=none] coordinates {(G-5,0.01)(Q-5,0.0)(Q-f,0.04)(L-5,0.01)(L-f,0.03)};

% --- Catalan ---
\nextgroupplot[
    title=Catalan,
    title style={yshift=-1ex},
    xticklabels={G-5, Q-5, Q-f, L-5, L-f},
    yticklabels={},
    enlarge x limits=0.2,
    bar width=11pt,
]
    \addplot+[ybar, fill=C1, draw=none] coordinates {(G-5,0.35)(Q-5,0.08)(Q-f,0.151)(L-5,0.04)(L-f,0.20)};
    \addplot+[ybar, fill=C2, draw=none] coordinates {(G-5,0.4)(Q-5,0.11)(Q-f,0.162)(L-5,0.05)(L-f,0.18)};
    \addplot+[ybar, fill=C3, draw=none] coordinates {(G-5,0.12)(Q-5,0.15)(Q-f,0.071)(L-5,0.06)(L-f,0.10)};
    \addplot+[ybar, fill=C4, draw=none] coordinates {(G-5,0.02)(Q-5,0.14)(Q-f,0.121)(L-5,0.33)(L-f,0.09)};
    \addplot+[ybar, fill=C5, draw=none] coordinates {(G-5,0.08)(Q-5,0.39)(Q-f,0.283)(L-5,0.46)(L-f,0.25)};
    \addplot+[ybar, fill=C6, draw=none] coordinates {(G-5,0.03)(Q-5,0.13)(Q-f,0.212)(L-5,0.06)(L-f,0.18)};

% ---  German ---
\nextgroupplot[
    title=German,
    title style={yshift=-1ex},
    xticklabels={G-5, Q-5, Q-f, L-5, L-f},
    yticklabels={},
    enlarge x limits=0.2,
    bar width=11pt,
]
    \addplot+[ybar, fill=C1, draw=none] coordinates {(G-5,0.48)(Q-5,0.26)(Q-f,0.175)(L-5,0.10)(L-f,0.212)};
    \addplot+[ybar, fill=C2, draw=none] coordinates {(G-5,0.27)(Q-5,0.07)(Q-f,0.072)(L-5,0.04)(L-f,0.040)};
    \addplot+[ybar, fill=C3, draw=none] coordinates {(G-5,0.08)(Q-5,0.15)(Q-f,0.144)(L-5,0.10)(L-f,0.101)};
    \addplot+[ybar, fill=C4, draw=none] coordinates {(G-5,0.01)(Q-5,0.08)(Q-f,0.083)(L-5,0.29)(L-f,0.091)};
    \addplot+[ybar, fill=C5, draw=none] coordinates {(G-5,0.15)(Q-5,0.42)(Q-f,0.433)(L-5,0.32)(L-f,0.404)};
    \addplot+[ybar, fill=C6, draw=none] coordinates {(G-5,0.01)(Q-5,0.02)(Q-f,0.093)(L-5,0.15)(L-f,0.152)};

% --- Japanese ---
\nextgroupplot[
    title=Japanese,
    title style={yshift=-1.5ex},
    xticklabels={G-5, Q-5, Q-f, L-5, L-f},
    yticklabels={},
    enlarge x limits=0.2,    
    bar width=11pt,
]
    \addplot+[ybar, fill=C1, draw=none] coordinates {(G-5,0.29)(Q-5,0.07)(Q-f,0.04)(L-5,0.03)(L-f,0.061)};
    \addplot+[ybar, fill=C2, draw=none] coordinates {(G-5,0.22)(Q-5,0.05)(Q-f,0.06)(L-5,0.02)(L-f,0.101)};
    \addplot+[ybar, fill=C3, draw=none] coordinates {(G-5,0.27)(Q-5,0.21)(Q-f,0.11)(L-5,0.11)(L-f,0.121)};
    \addplot+[ybar, fill=C4, draw=none] coordinates {(G-5,0.02)(Q-5,0.09)(Q-f,0.06)(L-5,0.2)(L-f,0.04)};
    \addplot+[ybar, fill=C5, draw=none] coordinates {(G-5,0.18)(Q-5,0.36)(Q-f,0.36)(L-5,0.33)(L-f,0.323)};
    \addplot+[ybar, fill=C6, draw=none] coordinates {(G-5,0.02)(Q-5,0.22)(Q-f,0.37)(L-5,0.31)(L-f,0.354)};

\end{groupplot}

\node[anchor=north] at ($(group c3r1.south) + (0, -0.7cm)$) {
  \begin{tikzpicture}
    \matrix[
      matrix of nodes,
      nodes={
        font=\normalsize,
        inner sep=2pt
      },
      column sep=2pt,
      row sep=2pt
    ] {
      \draw[fill=C1, draw=none, baseline] (0,0) rectangle (0.3cm,0.3cm); & \texttt{ACC} &
      \draw[fill=C2, draw=none, baseline] (0,0) rectangle (0.3cm,0.3cm); & \texttt{POT} &
      \draw[fill=C3, draw=none, baseline] (0,0) rectangle (0.3cm,0.3cm); & \texttt{Good} &
      \draw[fill=C4, draw=none, baseline] (0,0) rectangle (0.3cm,0.3cm); & \texttt{Unchanged} &
      \draw[fill=C5, draw=none, baseline] (0,0) rectangle (0.3cm,0.3cm); & \texttt{Degraded} &
      \draw[fill=C6, draw=none, baseline] (0,0) rectangle (0.3cm,0.3cm); & \texttt{Gibberish} \\
    };
    \draw[black, ultra thick] (-5.2, -0.60) -- (-5.1, -0.75) -- (-1.66, -0.75) -- (-1.56, -0.60);
    \draw[black, ultra thick] (0.94, -0.60) -- (1.04, -0.75) -- (4.88, -0.75) -- (4.98, -0.60);
    \node[below, black] at (-3.38, -0.82) {\texttt{Beneficial}};
    \node[below, black] at (2.96, -0.82) {\texttt{Harmful}};
  \end{tikzpicture}
};

\end{tikzpicture}
\caption{Distribution of output alternative categories. G: Gemma, Q: Qwen, L: Llama. -5: 5-shot, -f: fine-tuned.}
\label{fig:bar-manual}
\end{figure*}

\section{Results}

\subsection{Automatic Evaluation}

\autoref{tab:result_multils} shows the automatic metric scores for our LS systems.
The results confirm our hypothesis that fine-tuning, as part of the knowledge distillation, improved the performance of small LLMs. For example, fine-tuned Llama achieved 0.370 ACC on English, significantly higher than the 5-shot score (0.202).
Similar gains were observed for both Llama and Qwen across most languages.

The fine-tuned Llama performed comparably to Qwen despite its smaller size, suggesting that the 1B model can approach 1.5B model in performance after training.
However, neither student models reached the teacher's level.

\autoref{tab:result_multils} also reports the latency (ms/token) for prompt reading (read) and output generation (pred).
Both student models showed substantially lower latency than the teacher model. On m6g.large, Llama's read latency (66 msec/token) was nearly 10 times faster than Gemma's (652 msec/token), with similar trends across environments.

\subsection{Manual Evaluation}

\autoref{fig:bar-manual} shows the distribution of alternative categories, as judged by human evaluators, across models, settings, and languages. Each stacked bar represents the proportion of output alternatives falling into the categories.

Under 5-shot settings, small LLMs, especially Llama for English and Spanish, produced a high proportion of \texttt{Unchanged} outputs, indicating safer but less proactive simplification behavior.
Fine-tuning reduced \texttt{Unchanged} and corresponding rise in \texttt{Beneficial} simplifications, reflecting a general improvement in LS capability.
However, fine-tuning also introduced a safety trade-off, as it increased the proportions of \texttt{Harmful} alternatives.

In contrast, such trade-off was not observed for German and Japanese. For these languages, performance remained low across both 5-shot and fine-tuned settings, with \texttt{Harmful} alternatives consistently dominating the results. This suggests a more fundamental challenge stemming from the inherent difficulty for current small LLMs to perform LS effectively in these languages.

\begin{table}[t]
    \small
    \setlength{\tabcolsep}{5pt}
    \centering
    \begin{tabular}{ccccccc}
     Lang & Model & Settings & $r_B$ & $r_H$ & AUC & $B_{H_{0.1}}$ \\
    \midrule
    \multirow{4}{*}{En} & \multirow{2}{*}{Qwen} & 5-shot & .63 & .30 & .679 & .41 \\
    & & fine-tuned & .63 & .27 & \textbf{.707} & \textbf{.46} \\
    %\cmidrule{2-7}
    \noalign{\vskip 0.5ex}
    & \multirow{2}{*}{Llama} & 5-shot & .30 & .28 & .510 & .12\\
    & & fine-tuned & .61 & .20 & \textbf{.797} & \textbf{.54} \\
    \midrule
    \multirow{4}{*}{Es} & \multirow{2}{*}{Qwen} & 5-shot & .51 & .19 & .737 & .46 \\
    & & fine-tuned & .63 & .19 & \textbf{.850} & \textbf{.61} \\
    %\cmidrule{2-7}
    \noalign{\vskip 0.5ex}
    & \multirow{2}{*}{Llama} & 5-shot & .09 & .12 & \textbf{.907} & .09\\
    & & fine-tued & .56 & .21 & .804 & \textbf{.50} \\
    \midrule
    \multirow{4}{*}{Ca} & \multirow{2}{*}{Qwen} & 5-shot & .34 & .52 & .735 & .18 \\
    & & fine-tuned & .38 & .49 & \textbf{.904} & \textbf{.34} \\
    %\cmidrule{2-7}
    \noalign{\vskip 0.5ex}
    & \multirow{2}{*}{Llama} & 5-shot & .15 & .52 & .614 & .03 \\
    & & fine-tuned & .46 & .42 & \textbf{.813} & \textbf{.36} \\
    \midrule
    \multirow{4}{*}{De} & \multirow{2}{*}{Qwen} & 5-shot & .41 & .38 & \textbf{.841} & \textbf{.34} \\
    & & fine-tuned & .38 & .51 & .721 & .16 \\
    %\cmidrule{2-7}
    \noalign{\vskip 0.5ex}
    & \multirow{2}{*}{Llama} & 5-shot & .19 & .40 & .730 & .11\\
    & & fine-tuned & .35 & .55 & \textbf{.737} & \textbf{.16} \\
    \midrule
    \multirow{4}{*}{Ja} & \multirow{2}{*}{Qwen} & 5-shot & .33 & .58 & \textbf{.807} & \textbf{.16} \\
    & & fine-tuned & .21 & .73 & .799 & .13 \\
    %\cmidrule{2-7}
    \noalign{\vskip 0.5ex}
    & \multirow{2}{*}{Llama} & 5-shot & .16 & .64 & .745 & .04\\
    & & fine-tuned & .28 & .67 & \textbf{.845} & \textbf{.19} \\
    \bottomrule
    \end{tabular}
    \caption{Evaluation of Filtering Strategy. $r_B$ and $r_H$ refer to the original rate of \texttt{Beneficial} and \texttt{Harmful} outputs.}
    \label{tab:filtering}
\end{table}

\subsection{Filtering Strategy}
\autoref{tab:filtering} presents the results of filtering strategy.
First, the AUC scores are notably high, especially under fine-tuned settings, suggesting that log-probability serves as an effective signal for detecting \textbf{Harmful} alternatives. Moreover, the fine-tuned models generally show higher AUC across model types and languages, which indicates that knowledge distillation enhances the quality of probability as a safety indicator.

The $B_{H_{0.1}}$ metric shows the practical value of this strategy.
For example, in Spanish, fine-tuned Qwen reduced \texttt{Harmful} rate from 19\% to 10\% with only a slight drop in \texttt{Beneficial} from 63\% to 61\%. $B_{H_{0.1}}$ also highlights the superiority of fine-tuning to 5-shot settings.

\begin{figure}[t]
    \centering
    \scalebox{0.48}{\input{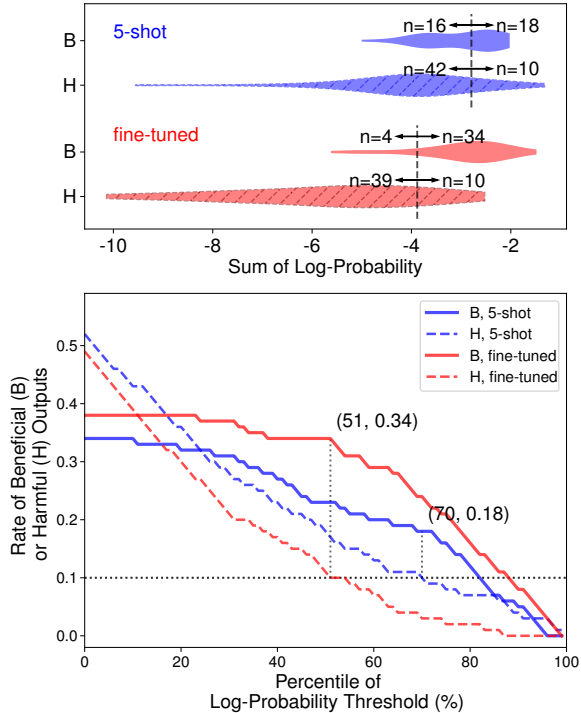}}
    \caption{\texttt{Beneficial} and \texttt{Harmful} alternatives and their probability of Qwen in Catalan. 
    (Top) Distribution of raw probability scores. (Bottom) Rate of \texttt{Beneficial} and \texttt{Harmful} alternatives after filtering at each percentile threshold.
    Dotted lines are plotted on thresholds where \texttt{Harmful} becomes 10\%.}
    \label{fig:filter}
\end{figure}

To further explore these findings, we focus on the behavior of Qwen models in Catalan.
Here, while the original \texttt{Beneficial} and \texttt{Harmful} rates are close between 5-shot and fine-tuned settings, the impact of filtering strategy differs significantly.
In \autoref{fig:filter}, the violin plot (top) visualizes the distribution of log-probability scores, where fine-tuning leads to a clear separation between \texttt{Beneficial} and \texttt{Harmful} alternatives.

The line plot (bottom) tracks \texttt{Beneficial} and \texttt{Harmful} rates across thresholds percentiles.
For the fine-tuned model, increasing the threshold reduces \texttt{Harmful} rapidly, while \texttt{Beneficial} declines more gradually.
As a result, the \texttt{Harmful} rate is reduced from nearly 50\% to 10\%, with most \texttt{Beneficial} simplification preserved.

\begin{table}[t]
    \small
    \centering
    \begin{tabular}{p{7cm}}
    \toprule

    \textbf{Context}: There are also different \underline{editing} styles in the sense of how bold people are willing to be.\\
    \textbf{Target Word}: editing\\
    \textbf{Gold Alternatives}: changing, modifying, altering ...\\
    \makebox[7cm][l]{\textbf{Gemma 5-shot} (4\%): writing \hfill (\textit{Change of Meaning})}\\
    \makebox[7cm][l]{\textbf{Qwen 5-shot} (92\%): writing \hfill (\textit{Change of Meaning})}\\
    \makebox[7cm][l]{\textbf{Qwen fine-tuned} (3\%): proofreading \hfill (\textit{More Difficult})}\\
    \bottomrule

    \end{tabular}
    \caption{Example outputs from the LS systems. Percentages next to system names indicate log-probability percentiles within each system.}
    \label{tab:case1}
\end{table}

\section{Discussion}

\subsection{Case Study}
To better understand the characteristics of model outputs, particularly harmful simplifications overlooked by automatic metrics and the potential of the log-probability signal, we present an example in \autoref{tab:case1}.
In this example, model output alternatives "writing" and "proofreading" were categorized as \texttt{Harmful}, with the tags "\textit{Change of Meaning}" and "\textit{More Difficult}", respectively.
Crucially, these alternatives were associated with lower log-probability percentiles for Gemma (5-shot) and fine-tuned Qwen, while they were much higher for Qwen under the 5-shot setting.
This case confirms our findings that fine-tuned models effectively leverage log-probability to identify harmful alternatives.
It also shows that log-probability is a useful signal for the teacher model, even without fine-tuning. This validates the filtering processed used during data synthesis. Examples in other languages are described in \autoref{sec:cases}.

\subsection{Safety}
\label{sec:safety}
As exemplified by the case study above, harmful LS alternatives pose a serious risk in real-world scenarios.
Our manual evaluation revealed key limitations of standard automatic evaluation metrics based on human-provided alternatives. They fail to identify acceptable simplifications not in the gold alternatives, and they do not expose harmful alternatives.
Although manual evaluation is costly and not scalable, our harmfulness annotations provide a valuable basis for building automatic detection methods, such as LLM-as-a-judge, to support more practical safety assessment.

Harmful alternatives were particularly pronounced in German and Japanese. In these languages, complex morphology may hinder the consistent generation of correct and simple single-word alternatives by small LLMs.
Our error analysis highlights a critical challenge related to this: alternatives with the \textit{Grammar Error} tag in German and Japanese often received high probability scores from small LLMs (both few-shot and fine-tuned), making them difficult to distinguish from beneficial alternatives or other harmful types.
For instance, the average log-probability score for \textit{Grammar Error} from the fine-tuned Llama model in Japanese was -2.992, which was notably higher than that for \textit{Change of Meaning} (-3.762) and \textit{Gibberish} (-4.457). This suggests that our filtering strategy had limited effectiveness in mitigating grammar errors.

Interestingly, this issue was less prevalent in the teacher model (see \autoref{sec:cats} for details across all tags and models).
This disparity implies that non-small LLMs can better leverage output probability as a signal for grammatical correctness even in morphologically complex languages. In contrast, small LLMs may struggle to capture these fine-grained grammatical nuances with simple approaches such as in-context learning and knowledge distillation. Incorporating instances with grammatical errors as negative examples in contrastive learning may help student models learn to avoid them, enhancing the reliability of threshold-based filtering.

While log-probability is effective for filtering harmful alternatives, selecting an appropriate threshold for real-world use requires careful tuning based on human evaluation, taking into account domain- and language-specific considerations and practical application needs, to ensure both safety and utility.

\subsection{Latency}
While the smaller models offer substantial speed improvements, their practical inference speed for real-time and on-device LS needs further consideration.
Assuming that a standard input consists of 30 tokens and the output alternative word is composed of two tokens, the overall inference time for fine-tuned Llama on the faster m6g.xlarge environment would be about 1.2 seconds: (30 tokens * 33 ms/token [read]) + (2 tokens * 107 ms/token [pred]) = 1204 ms.

Although a response time of around one second may be tolerable in some cases, further reduction would likely improve the user experience on mobile devices. One possible approach is to reduce the prompt size by including only a limited window of words surrounding the target, rather than the full sentence. Naturally, this strategy would require careful safety assessment.

\section{Conclusion}

This study addressed the challenge of building efficient and safe LS systems using small LLMs, motivated by real-world needs. We proposed benchmark systems in five languages based on in-context learning and knowledge distillation, and introduced a filtering strategy using log-probability as a safety signal.
Experiments showed that small LLMs offer significant efficiency gains, but that knowledge distillation, while improving automatic metrics score, increases harmful outputs.

We demonstrated that output log-probability serves as an effective signal for detecting harmful simplifications. This signal enables filtering strategy that reduce harmful outputs while retaining beneficial ones. These findings lay the foundation for lightweight LS systems that remain safe and effective across languages.

Future work should improve training methods to reduce harmfulness and explore real-time LS for mobile environments. Ultimately, this research advances deployable, trustworthy LS tools that support inclusive information access.

\section*{Limitations}

Our study, while demonstrating the potential of small LLMs for efficient and safer lexical simplification, has several limitations that highlight directions for further investigation.

First, the manual evaluation of harmfulness was conducted by a single annotator per language. While the annotation task was designed as a simple binary decision to minimize subjectivity, we were unable to assess inter-annotator agreement, which may affect the generalizability of the harmfulness evaluations. Establishing a more robust evaluation protocol with multiple annotators would be a valuable next steop to create a gold-standard dataset for harmfulness detection in LS.

Next, we employed relatively simple prompt engineering, using fixed 5-shot examples and prompt templates to ensure reproducibility and establish baseline performance. We did not explore advanced prompt engineering techniques, which could potentially enhance the models' performance. Future work could investigate how more sophisticated prompting strategies impact the trade-off between performance and safety explored in this study.

This study adopted a narrow task definition, focusing on generating a single simpler alternative for each target word. The systems were not designed to produce multiple candidate simplifications or to handle multi-word expressions, which are often important for user understanding and for simplifying nuanced concepts. Extending our framework to sentence- or paragpraph-level simplification would be a crucial step towards more practical TS tools.

Furthermore, our investigation focused only on generating simpler alternatives. Other important aspects of lexical simplification, such as identifying complex words and selecting the most appropriate simplification, were not addressed in this work. Integrating our safety-aware models into a full LS pipeline is an essential direction for future research.

Finally, the sensitivity of model performance to quantization is a critical limitation. Our methodology involves distinct quantization steps, 4-bit precision during fine-tuning and GGUF for deployment, which can introduce performance discrepancies. Although we observed only negligible performance changes in this study, smaller models are generally more vulnerable to degradation from such processes. Therefore, there is a possibility that our framework might not operate as expected under different quantization schemes, potentially affecting its reliability.
\section*{Ethical Considerations}

We used publicly available data sources and followed their respective licenses. The training data is from Wikipedia, which is available under the Creative Commons Attribution-ShareAlike 3.0 Unported (CC BY-SA 3.0) license. For evaluation, we used the MultiLS dataset, also distributed under a CC BY-SA license. The code used for training and evaluation will be made publicly available upon publication.

\section*{Acknowledgments}
This work is partially financed by the Ministerio de Ciencia, Innovación y Universidades, Agencia Estatal de Investigaciones: project CPP2023-010780 funded by MICIU/AEI/10.13039/501100011033 and by FEDER, UE (“Habilitando Modelos de Lenguaje Responsables e Inclusivos”). We also acknowledge funding from the European Union’s Horizon Europe research and innovation program under Grant Agreement No. 101132431 (iDEM Project).  Horacio Saggion also receives support from the Spanish State Research Agency under the Maria de Maeztu Units of Excellence Programme (CEX2021-001195-M) and from the Departament de Recerca i Universitats de la Generalitat de Catalunya (ajuts SGR-Cat 2021).

% Bibliography entries for the entire Anthology, followed by custom entries
\bibliography{anthology,custom}

\clearpage
\appendix
\section{Prompts Provided to LLMs}
\label{sec:prompts}

We provided the prompt in \autoref{tab:prompt_5shot} for few-shot learning and in \autoref{tab:prompt_trained} for fine-tued models. The prompt for fine-tuning was shortened to minimize inference time.

\begin{table}[t]
    \small
    \centering
    \begin{tabular}{p{7cm}}
    \toprule
    \texttt{Given the context and the specified target in} \textit{\{language\}}\texttt{, provide a simpler alternative word.}\\
    \\
    \textit{\{5-shot examples\}}\\
    \\
    \texttt{Context:} \textit{\{context\}}\\
    \texttt{Target Word:} \textit{\{target\}}\\
    \texttt{Alternative Word:}\\
    \bottomrule
    \end{tabular}
    \caption{Prompt template for 5-shot settings.}
    \label{tab:prompt_5shot}
\end{table}

\begin{table}[t]
    \small
    \centering
    \begin{tabular}{p{7cm}}
    \toprule    
    \texttt{Context:} \textit{\{context\}}\\
    \texttt{Target Word:} \textit{\{target\}}\\
    \texttt{Simplified:} \textit{\{alternative\}}\\
    \bottomrule
    \end{tabular}
    \caption{Prompt template for training models. \textit{\{alternative\}} was removed during the inference of fine-tuned models.}
    \label{tab:prompt_trained}
\end{table}

\section{Hyperparameters}
\label{sec:hps}

\autoref{tab:hps} shows the hyperparameter settings we used for inference and training. For other hyperparameters, we used default values of \texttt{GenerationConfig}, \texttt{TrainingArguments}, and \texttt{LoraConfig} classes of Huggingface Transformers.

\begin{table}[t]
    \centering
    \begin{tabular}{cc}
    \toprule
    \multicolumn{2}{c}{\textbf{Inference}} \\
    \midrule
    Parameter & Value \\
    \midrule
    Decoding & Greedy \\
    Sampling & Disabled \\
    Temperature & 1.0 \\
    Max generation length & 10\\
    \midrule    
    \multicolumn{2}{c}{\textbf{Training}} \\
    \midrule
    Parameter & Value \\
    \midrule
    Optimizer & AdamW \\
    Weight decay & 0.01 \\
    Learning Rate & 3e-5 \\
    Scheduler & Linear \\
    Batch Size & 16 \\
    Max Epoch & 5 \\
    Lora r & 8 \\
    Lora alpha & 4 \\
    Lora dropout & 0.1 \\
    \bottomrule
    \end{tabular}
    \caption{Hyperparameters for training and inference.}
    \label{tab:hps}
\end{table}

\section{Model Selection}
\label{sec:it-models}

For selecting the teacher model, we considered LLMs that are mid-sized, open-source, and capable of high multilingual performance, taking into account the need to synthesize large amounts of data.
\autoref{tab:result_multils_dev} presents the results for the candidate models: Qwen 2.5 14B, Phi-3 medium \citep{abdin2024phi3technicalreporthighly}, and Gemma 2 9B.
Although these models did not reach the performance of state-of-the-art LS system by \citet{enomoto-etal-2024-tmu}, which used GPT-4 along with ensembling and reranking, Gemma 2 9B was selected due to its relatively small size and balanced high performance across languages.

For selecting suitable lightweight models, we initially considered Gemma 2 2B, Llama 3.2 1B, Qwen 2.5 1.5B, and Qwen 2.5 0.5B due to their multilingual support and small model size.
Gemma 2 2B was excluded due to its latency on the m6g.xlarge instance, where the base model with 5-shot setting required 139 ms/token for reading and 476 ms/token for prediction, which was not sufficient for practical use.
For the remaining LLMs, we evaluated performance on development set across both the base and instruction-tuned models under three settings: 0-shot, 5-shot, and knowledge distillation. In the 0-shot setting, the prompt was created by removing \textit{\{5-shot examples\}} from the prompt in \autoref{tab:prompt_5shot}.

In the results in \autoref{tab:result_multils_dev}, the following trends were observed. Firstly, Qwen 2.5 0.5B consistently showed poor performance across all settings. Next, for other models, the 5-shot setting generally outperformed 0-shot. Lastly, while instruction-tuned models slightly outperformed base models in the 5-shot setting, the base models achieved better performance in the knowledge distillation setting. Based on these results, we selected Qwen 2.5 1.5B and Llama 3.2 1B as representative models. To ensure a fair comparison of the proposed methods, we used the base models for both 5-shot and knowledge distillation settings.

\begin{table*}[t]
    \centering
    \setlength{\tabcolsep}{4pt}
    \small
    \begin{tabular}{ccccccccccccc}
        \toprule
        \multirow{2}{*}{Model} & \multirow{2}{*}{IT} & \multirow{2}{*}{Settings} & \multicolumn{2}{c}{English} & \multicolumn{2}{c}{Spanish} & \multicolumn{2}{c}{Catalan} & \multicolumn{2}{c}{German} & \multicolumn{2}{c}{Japanese} \\
         & & & ACC & POT & ACC & POT & ACC & POT & ACC & POT & ACC & POT \\
         \midrule
        %\hline
        GPT-4 & - & - & .522 & .833 & .578 & .844 & .489 & .767 & .544 & .800 & .478 & .722 \\
        Qwen 2.5 14B & \checkmark & 5-shot & .511 & .767 & .444 & .778 & .289 & .600 & .400 & .611 & .244 & .489 \\
        Phi 3 medium (14B) & \checkmark & 5-shot & .467 & .733 & .478 & .733 & .200 & .467 & .367 & .567 & .278 & .433 \\
        Gemma 2 9B & \checkmark & 5-shot & .489 & .700 & .422 & .711 & .333 & .611 & .478 & .689 & .200 & .444 \\
        \midrule
         & & 0-shot & \textbf{.311} & .467 & .089 & .167 & .044 & .111 & .000 & .056 & .067 & .122 \\
         & \checkmark & 0-shot & .289 & \textbf{.489} & \underline{\textbf{.333}} & \underline{\textbf{.544}} & \textbf{.067} & \textbf{.200} & \textbf{.111} & \textbf{.200} & \textbf{.067} & \underline{\textbf{.178}} \\
        %\noalign{\vskip 0.5ex}         
        \multirow{2}{*}{Qwen 2.5 1.5B} & & 5-shot & .300 & .522 & .300 & \textbf{.511} & .067 & \underline{\textbf{.244}} & \textcolor{red}{\underline{{\textbf{.178}}}} & \textcolor{red}{\underline{\textbf{.289}}} & \textcolor{red}{\underline{\textbf{.089}}} & .144 \\
         & \checkmark & 5-shot & \underline{\textbf{.333}} & \underline{\textbf{.533}} & \textbf{.322} & .500 & \underline{\textbf{.078}} & \underline{\textbf{.244}} & .144 & .256 & \textcolor{red}{\underline{\textbf{.089}}} & \underline{\textbf{.178}} \\
        %\cmidrule{2-7}
        \noalign{\vskip 1ex}         
         & & fine-tuned & \textcolor{red}{\textbf{.344}} & .533 & \textcolor{red}{\textbf{.378}} & \textcolor{red}{\textbf{.611}} & \textcolor{red}{\textbf{.144}} & \textcolor{red}{\textbf{.256}} & \textbf{.144} & \textbf{.267} & \textcolor{red}{\textbf{.089}} & \textcolor{red}{\textbf{.211}} \\
         & \checkmark & fine-tuned & \textcolor{red}{\textbf{.344}} & \textcolor{red}{\textbf{.567}} & .278 & .433 & .022 & .133 & .089 & .144 & .033 & .111 \\
        \midrule
         & & 0-shot & .022 & .022 & .000 & .044 & .000 & .011 & .000 & .000 & .011 & .011 \\
        & \checkmark & 0-shot & \textbf{.211} & \textbf{.378} & \textbf{.078} & \textbf{.189} & .000 & \textbf{.022} & \textbf{.056} & \textbf{.089} & \textbf{.044} & \textbf{.078} \\
        %\noalign{\vskip 0.5ex}
        \multirow{2}{*}{Llama 3.2 1B} & & 5-shot & .211 & .300 & .022 & .078 & \underline{\textbf{.033}} & \underline{\textbf{.144}} & \underline{\textbf{.089}} & \underline{\textbf{.122}} & \underline{\textbf{.056}} & .078 \\
         & \checkmark & 5-shot & \underline{\textbf{.289}} & \underline{\textbf{.533}} & \underline{\textbf{.233}} & \underline{\textbf{.356}} & \underline{\textbf{.033}} & .122 & \underline{\textbf{.089}} & \underline{\textbf{.122}} & \underline{\textbf{.056}} & \underline{\textbf{.111}} \\
        \noalign{\vskip 1ex}         
         & & fine-tuned & \textcolor{red}{\textbf{.444}} & \textcolor{red}{\textbf{.622}} & \textcolor{red}{\textbf{.367}} & \textcolor{red}{\textbf{.544}} & \textcolor{red}{\textbf{.167}} & .289 & \textcolor{red}{\textbf{.189}} & .244 & \textcolor{red}{\textbf{.122}} & \textcolor{red}{\textbf{.200}} \\
         & \checkmark & fine-tuned & .422 & \textcolor{red}{\textbf{.622}} & .267 & .478 & .122 & \textcolor{red}{\textbf{.333}} & .156 & \textcolor{red}{\textbf{.256}} & .022 & .156 \\
        \midrule
         & & 0-shot & .144 & .178 & .056 & .111 & \underline{\textbf{.011}} & \underline{\textbf{.044}} & \underline{\textbf{.011}} & \textbf{.033} & \textbf{.022} & \textbf{.056} \\
        & \checkmark & 0-shot & \underline{\textbf{.156}} & \textbf{.233} & \underline{\textbf{.111}} & \underline{\textbf{.233}} & \underline{\textbf{.011}} & \underline{\textbf{.044}} & \underline{\textbf{.011}} & \textbf{.033} & .000 & .011 \\
        \multirow{2}{*}{Qwen 2.5 0.5B} & & 5-shot & .033 & .067 & .022 & .056 & \underline{\textbf{.011}} & \textbf{.011} & \underline{\textbf{.011}} & \underline{\textbf{.044}} & .033 & \underline{\textbf{.067}} \\
         & \checkmark & 5-shot & \textbf{.144} & \underline{\textbf{.244}} & \textbf{.089} & \textbf{.133} & .000 & \textbf{.011} & \underline{\textbf{.011}} & .022 & \underline{\textbf{.044}} & \underline{\textbf{.067}}\\
        \noalign{\vskip 1ex}
         & & fine-tuned & .200 & .344 & \textcolor{red}{\textbf{.189}} & \textcolor{red}{\textbf{.311}} & \textcolor{red}{\textbf{.033}} & \textcolor{red}{\textbf{.067}} & \textcolor{red}{\textbf{.044}} & \textcolor{red}{\textbf{.056}} & \textcolor{red}{\textbf{.067}} & \textcolor{red}{\textbf{.111}} \\
         & \checkmark & fine-tuned & \textcolor{red}{\textbf{.267}} & \textcolor{red}{\textbf{.389}} & .156 & .256 & .000 & .022 & .022 & .022 & .056 & \textcolor{red}{\textbf{.111}} \\
        \bottomrule
    \end{tabular}
    \caption{Performance on MultiLS across various models and settings. For the performance of GPT-4, we used outputs of \citet{enomoto-etal-2024-tmu}. Checkmarks on the IT column refer to the performance from instruction-tuned version. \textbf{Bold} numbers are the better scores between the base and instruction-tuned models under the same setting. \underline{Underlined} numbers are the best performance among 0-shot and 5-shot settings. \textcolor{red}{Red} numbers are the best performance across all settings.}
    \label{tab:result_multils_dev}
\end{table*}

\section{Japanese Tokenization}
\label{sec:ja-tokens}

Since Japanese does not use spaces to separate words, tokenization is required to extract individual words. We primarily used MeCab for tokenization. However, considering the characteristics of the target words in MultiLS, we applied the following rules to select candidate words during data syntheisis: (1) Consecutive nouns were grouped together as a single unit; (2) For inflected parts-of-speech such as verbs and adjectives, auxiliary verbs were included along with the word stem. 
It should be noted that the above rules may not always yield exact matches, as the dataset includes multi-word expressions as target words.

\section{Manual Evaluation Examples}
\label{sec:tags}

\begin{table}[t]
    \small
    \centering
    \begin{tabular}{c|cccc|c}
    \toprule
    \multicolumn{6}{p{7cm}}{\textbf{Context}: An \underline{ingenious} alphabet allowed the Maya to record information on their monuments and temples, giving anthropologists an excellent way to learn about Maya life and culture.} \\
    \multicolumn{6}{l}{\textbf{Target Word}: ingenious}\\
    \midrule

    Alternative & \textit{GE} & \textit{CM} & \textit{MD} & \textit{GB} & Group\\
    \midrule
    innovative & & & & & \texttt{Good} \\
    innovatively & \checkmark & & & & \texttt{Degraded} \\
    sophisticated & & & & & \texttt{Good} \\
    adroit & & & \checkmark & & \texttt{Degraded} \\
    anonymous & & \checkmark & & & \texttt{Degraded} \\
    anonymously & & & & \checkmark & \texttt{Gibberish} \\
     simple & & \checkmark & & & \texttt{Degraded} \\
    simply & \checkmark & \checkmark & & & \texttt{Degraded} \\
    \bottomrule
    \end{tabular}
    \caption{Example tags provided to annotators.}
    \label{tab:ex-tags}
\end{table}

\autoref{tab:ex-tags} shows examples of harmfulness tags assigned to alternatives. These are provided to annotators as reference.

In this example, "ingenious" is the target word to be simplified. While "sophisticated" and "innovative" are appropriate simplifications, other alternatives are harmful. Although replacing "ingenious" with "sophisticated" makes the sentence ungrammatical due to the article-adjective agreement (an sophisticated), such an inconsistency is not considered a harmful simplification in our evaluation.

\section{Simplification Examples}
\label{sec:cases}

\autoref{tab:cases_appendix} presents examples for languages other than English. 

In the Spanish example, the target word "desequilibrado" \textit{(not in equilibrium)} was simplified to "equilibrado" \textit{(in equilibrium)} by Llama under both 5-shot and fine-tuned settings, which reversed the meaning of the context. These harmful outputs had high log-probability scores, which made them difficult to eliminate through the filtering strategy. On the other hand, the teacher model produced a beneficial output, but its low log-probability would likely lead to its removal.

In the Catalan example, fine-tuned Llama created an adverb-looking word combining the word "mal" \textit{(bad)} with a replication of adverbial suffixes "-ment". This output is clearly \textit{Gibberish}, and similar cases were observed multiple times in the fine-tuned model. Such outputs need to be removed, and the filtering strategy is likely to be effective in achieving this.

In the German example, the output from Gemma 5-shot and Llama 5-shot were assigned \textit{Grammar Error}, while the output from Llama fine-tuned was assigned \textit{More Difficult}. 
For Llama 5-shot, a noun was proposed while the output should be an adjective as with the target word. This suggests that the system failed to fully understand the task of providing a contextually appropriate word. In German, capitalized words indicate nouns. However, due to the auto-regressive nature of the output, previously generated tokens cannot be revised. General methods such as beam search can mitigate this issue, but they are not applicable to real-time generation, and thus solutions will rely on strategies during training.
For the teacher model, grammatical agreement requires "grundlegender" rather than the output "grundlegende". The output is nearly correct, and a finer-grained language-specific tags may be needed for further analysis.
The output from Llama fine-tuned fits the and preserves the intended meaning, but the word appears to be an invented term. \textit{More Difficult} was assigned to this output, and its low log-probability suggests that this kind of words could be filtered out.

Lastly, in the Japanese example, both outputs from Qwen were assigned \textit{Grammar Error}. Both systems attempted to produce the appropriate verb "\begin{CJK}{UTF8}{min}使う\end{CJK}", but the Qwen 5-shot output contains an incorrect inflection, while the Qwen fine-tuned output lacks an inflectional suffix. These outputs have relatively high log-probabilities, and therefore it is difficult to filter them out.

\begin{table*}[t]
    \small
    \centering
    \begin{tabular}{p{13cm}}
    \toprule
    \textbf{Spanish} \\
    \midrule
    \textbf{Context}: Pero si eso ocurre habitualmente, tienes un flujo de fondos negativo y tu presupuesto está \underline{desequilibrado}.\\
    \textit{(But if that happens habitually, you have a negative cash flow and your budget is \underline{not in equilibrium}.)} \\
    \textbf{Target Word}: desequilibrado \textit{(not in equilibrium)}\\
    \textbf{Gold Alternatives}: inestable \textit{(unstable)}, desnivelado \textit{(uneven)}, desbalanceado \textit{(unbalanced)} ...\\
    \makebox[13cm][l]{\textbf{Gemma 5-shot} (5\%): desbalanceado \textit{(unbalanced)} \hfill (\texttt{Beneficial (POT)})}\\
    \makebox[13cm][l]{\textbf{Llama 5-shot} (55\%): equilibrado \textit{(in equilibrium)} \hfill (\textit{Change of Meaning})}\\
    \makebox[13cm][l]{\textbf{Llama fine-tuned} (77\%): equilibrado \textit{(in equilibrium)} \hfill (\textit{Change of Meaning})}\\
    \midrule

    \textbf{Catalan} \\
    \midrule
    \textbf{Context}: En el manifest s'ha qualificat "d'escandalosa" la sentència contra els membres de "la Manada" ja que "se'n riu i menysprea una dona jove" que va ser agredia "\underline{brutalment} per un grup de salvatges".\\
    \textit{(In the statement, the sentence against the members of "la Manada" was described as "scandalous" since "laughs at and despises a young woman" who was assaulted "\underline{bruttally} by a group of savages".)} \\
    \textbf{Target Word}: brutalment \textit{(bruttally)} \\
    \textbf{Gold Alternatives}: violentament \textit{(violently)}, fortament \textit{(strongly)}, durament \textit{(severely)} ...\\
    \makebox[13cm][l]{\textbf{Gemma 5-shot} (51\%): violentament \textit{(violently)} \hfill (\texttt{Beneficial(POT)})}\\
    \makebox[13cm][l]{\textbf{Llama 5-shot} (41\%): brutalment \textit{(bruttally)} \hfill (\texttt{Unchanged})}\\
    \makebox[13cm][l]{\textbf{Llama fine-tuned} (8\%): malamentamentamentamentamentamentamentamentament \hfill (\textit{Gibberish})}\\
    \midrule

    \textbf{German} \\
    \midrule
    \textbf{Context}: Salzborn nennt als in die moderne Begriffsgenese von Demokratie eingeschriebene Werte: (...) und die Gewähr \underline{elementarer} Rechte der Menschen gegen den Staat.\\
    \textit{(Salzborn names as values inscribed into the modern conceptual genesis of democracy: (...) and the guarantee of \underline{elementary} rights of human beings against the state.)} \\
    \textbf{Target Word}: elementarer \textit{(elementary)}\\
    \textbf{Gold Alternatives}: grundlegender \textit{(fundamental)}, wichtiger \textit{(important)}, essentieller \textit{(essential)} ...\\
    \makebox[13cm][l]{\textbf{Gemma 5-shot} (57\%): grundlegende \textit{(fundamental)} \hfill (\textit{Grammar Error})}\\
    \makebox[13cm][l]{\textbf{Llama 5-shot} (35\%): Grundrecht \textit{(fundamental right)}\hfill (\textit{Grammar Error})} \\
    \makebox[13cm][l]{\textbf{Llama fine-tuned} (9\%): grundstehender \textit{(ground-standing)}\hfill (\textit{More Difficult})}\\
    \midrule

    \textbf{Japanese} \\
    \midrule
    \textbf{Context}: \begin{CJK}{UTF8}{min}迅速に適切な解決を図るために、相談窓口を\underline{活用する}ことをお奨めします。\end{CJK}\\
    \textit{(To ensure a prompt and appropriate resolution, we recommend \underline{utilizing} the consulation service.)} \\
    \textbf{Target Word}: \begin{CJK}{UTF8}{min}活用する\end{CJK} \textit{(utilizing)}\\
    \textbf{Gold Alternatives}: \begin{CJK}{UTF8}{min}使う\end{CJK} \textit{(use)}, \begin{CJK}{UTF8}{min}利用する\end{CJK} \textit{(make use of)}, ...\\
    \makebox[13cm][l]{\textbf{Gemma 5-shot} (97\%): \begin{CJK}{UTF8}{min}利用する\end{CJK} \textit{(make use of)} \hfill (\texttt{Beneficial (ACC)})}\\
    \makebox[13cm][l]{\textbf{Qwen 5-shot} (63\%): \begin{CJK}{UTF8}{min}使おう\end{CJK} \hfill (\textit{Grammar Error})}\\
    \makebox[13cm][l]{\textbf{Qwen fine-tuned} (76\%): \begin{CJK}{UTF8}{min}使\end{CJK} \hfill (\textit{Grammar Error})}\\
    \bottomrule
    \end{tabular}
    \caption{Example outputs from the LS systems. Percentages next to system names indicate log-probability percentiles within each system.}
    \label{tab:cases_appendix}
\end{table*}

\section{Probability Scores across Categories}
\label{sec:cats}

\autoref{tab:error_cats} shows the distribution of harmful tags and their average log-probabilities for each model and language.
\textit{More Difficult} is generally rare, but the distribution of other tags varies across models and languages. As mentioned in \autoref{sec:safety}, the average log-probability score of \textit{Grammar Error} from small LLMs in German and Japanese are  higher, often comparable or sometimes superior to that of entire outputs. This trend is not pronounced in other languages or from the teacher model.

\begin{table*}[t]
    \small
    \centering
    \begin{tabular}{lcccccccccc}
    \toprule
     & \multicolumn{2}{c}{English} & \multicolumn{2}{c}{Spanish} & \multicolumn{2}{c}{Catalan} & \multicolumn{2}{c}{German} & \multicolumn{2}{c}{Japanese} \\
     Tags & \# & Logprob & \# & Logprob & \# & Logprob  & \# & Logprob  & \# & Logprob \\
    \midrule
    \multicolumn{11}{c}{\textbf{Gemma-5shot}} \\
    \midrule
     (All) & 100 & -1.615 & 100 & -1.567 & 100 & -1.679 & 100 & -1.588 & 100 & -2.268 \\
     \textit{More Difficult} & 4 & -1.905 & 2 & -1.989 & 0 & - & 1 & -1.300 & 4 & -2.031 \\
     \textit{Change of Meaning} & 14 & -1.874 & 2 & -1.266 & 6 & -1.907 & 6 & -1.620 & 4 & -2.447 \\
     \textit{Grammar Error} & 1 & -2.158 & 2 & -1.409 & 3 & -1.836 & 7 & -1.835 & 10 & -2.528 \\
     \textit{Gibberish} & 3 & -2.056 & 1 & -2.944 & 3 & -2.227 & 1 & -1.975 & 2 & -3.617 \\
    \midrule
    \multicolumn{11}{c}{\textbf{Qwen-5shot}} \\
    \midrule
     (All) & 100 & -1.884 & 100 & -2.129 & 100 & -3.592 & 100 & -2.754 & 100 & -4.132 \\
     \textit{More Difficult} & 2 & -2.203 & 1 & -3.013 & 0 & - & 3 & -3.217 & 3 & -3.882 \\
     \textit{Change of Meaning} & 20 & -2.088 & 8 & -2.957 & 24 & -3.976 & 20 & -3.834 & 23 & -4.766 \\
     \textit{Grammar Error} & 3 & -2.339 & 12 & -2.469 & 24 & -3.927 & 20 & -3.254 & 15 & -4.220 \\
     \textit{Gibberish} & 5 & -1.991 & 0 & - & 13 & -4.440 & 2 & -4.253 & 2 & -5.209 \\
    \midrule
    \multicolumn{11}{c}{\textbf{Qwen-fine-tuned}} \\
    \midrule
     (All) & 100 & -1.297 & 100 & -2.063 & 100 & -4.431 & 100 & -3.667 & 100 & -3.337 \\
     \textit{More Difficult} & 2 & -2.033 & 0 & - & 0 & - & 1 & -4.934 & 1 & -3.421 \\
     \textit{Change of Meaning} & 14 & -1.617 & 12 & -3.001 & 18 & -5.692 & 34 & -4.250 & 21 & -3.697 \\
     \textit{Grammar Error} & 5 & -1.514 & 9 & -3.390 & 17 & -5.161 & 10 & -3.296 & 17 & -3.018 \\
     \textit{Gibberish} & 7 & -1.408 & 4 & -5.029 & 21 & -6.047 & 9 & -5.206 & 37 & -4.021 \\
    \midrule
    \multicolumn{11}{c}{\textbf{Llama-5shot}} \\
    \midrule
     (All) & 100 & -1.807 & 100 & -1.244 & 100 & -2.873 & 100 & -3.135 & 100 & -4.204 \\
     \textit{More Difficult} & 3 & -1.603 & 1 & -1.802 & 0 & - & 2 & -4.501 & 0 & - \\
     \textit{Change of Meaning} & 14 & -2.045 & 8 & -1.573 & 32 & -3.246 & 13 & -3.537 & 16 & -4.520 \\
     \textit{Grammar Error} & 0 & - & 3 & -1.851 & 28 & -3.374 & 14 & -3.275 & 19 & -3.011 \\
     \textit{Gibberish} & 12 & -1.604 & 1 & -2.686 & 6 & -3.517 & 13 & -3.375 & 31 & -6.016 \\
    \midrule
    \multicolumn{11}{c}{\textbf{Llama-fine-tuned}} \\
    \midrule
     (All) & 100 & -1.161 & 100 & -1.862 & 100 & -2.880 & 100 & -3.645 & 100 & -3.360 \\
     \textit{More Difficult} & 0 & - & 0 & - & 0 & - & 4 & -4.867 & 3 & -4.091 \\
     \textit{Change of Meaning} & 11 & -1.465 & 16 & -2.720 & 17 & -3.012 & 25 & -4.147 & 20 & -3.762 \\
     \textit{Grammar Error} & 3 & -1.764 & 6 & -2.834 & 13 & -3.260 & 14 & -3.304 & 12 & -2.992 \\
     \textit{Gibberish} & 6 & -1.697 & 3 & -2.219 & 18 & -4.415 & 15 & -4.918 & 35 & -4.457 \\
     \bottomrule
    \end{tabular}
    \caption{Average log-probability scores for each language and harmful tag.}
    \label{tab:error_cats}
\end{table*}

\end{document}